\documentclass[11pt]{article}

\usepackage[preprint]{arr/acl}

\usepackage{times}
\usepackage{latexsym}

\usepackage[T1]{fontenc}

\usepackage[utf8]{inputenc}

\usepackage{microtype}

\usepackage{inconsolata}
\usepackage[table]{xcolor}
\usepackage{amsmath,amssymb}
\usepackage{booktabs}
\usepackage{adjustbox}
\usepackage{makecell}

\usepackage{graphicx}

\newcommand{\methodname}{\textsc{FrameSkip}}

\title{FrameSkip: Learning from Fewer but More Informative Frames \\ in VLA Training}

\author{
 \textbf{Bin Yu\textsuperscript{1,2,\thanks{Equal Contribution}}},
 \textbf{Shijie Lian\textsuperscript{2,4,\footnotemark[1]}},
 \textbf{Xiaopeng Lin\textsuperscript{3,6,\footnotemark[1]}},
 \textbf{Zhaolong Shen\textsuperscript{2,7,\footnotemark[1]}},
 \textbf{Yuliang Wei\textsuperscript{1,\thanks{Corresponding author}}},
\\
 \textbf{Changti Wu\textsuperscript{2,5}},
 \textbf{Hang Yuan\textsuperscript{2,5}},
 \textbf{Haishan Liu\textsuperscript{2}},
 \textbf{Bailing Wang\textsuperscript{1}},
 \textbf{Cong Huang\textsuperscript{2,3}},
 \textbf{Kai Chen \textsuperscript{2,3,8,\footnotemark[2]}}
\\
\\
 {\small \textsuperscript{1}Harbin Institute of Technology,
 \textsuperscript{2}Zhongguancun Academy}
\\
 {\small \textsuperscript{3}Zhongguancun Institute of Artificial Intelligence}
\\
 {\small \textsuperscript{4}Huazhong University of Science and Technology,
 \textsuperscript{5}East China Normal University}
\\
 {\small \textsuperscript{6}The Hong Kong University of Science and Technology (Guangzhou),
 \textsuperscript{7}Beihang University}
\\
 {\small \textsuperscript{8}DeepCybo}
}

\begin{document}
\maketitle
\footnotetext[3]{Work done at Zhongguancun Academy (Beijing).}

\begin{abstract}
Vision-Language-Action (VLA) policies are commonly trained from dense robot demonstration trajectories, often collected through teleoperation, by sampling every recorded frame as if it provided equally useful supervision.
We argue that this convention creates a temporal supervision imbalance: long low-change segments dominate the training stream, while manipulation-critical transitions such as alignment, contact, grasping, and release appear only sparsely.
We introduce \textbf{\methodname{}}, a data-layer frame selection framework that scores trajectory frames using action variation, visual-action coherence, task-progress priors, and gripper-transition preservation, then remaps training samples toward high-importance frames under a target retention ratio.
Because \methodname{} operates only in the dataloader, it leaves the VLA architecture, action head, training objective, and inference procedure unchanged.
Across RoboCasa-GR1, SimplerEnv, and LIBERO, \methodname{} improves the success-retention trade-off over full-frame training and simpler frame selection variants, achieving a macro-average success rate of 76.15\% across the three benchmarks compared with 66.50\% for full-frame training while using a compressed trajectory view that retains 20\% of unique frames in the main setting.
Code and model checkpoints are available on \href{https://github.com/ZGC-EmbodyAI/FrameSkip}{GitHub} and \href{https://huggingface.co/collections/VLyb/frameskip}{Hugging Face}.
\end{abstract}

\providecommand{\methodname}{\textsc{FrameSkip}}

\section{Introduction}
\label{sec:intro}

Vision-Language-Action (VLA) models have recently emerged as a promising paradigm for robotic manipulation by combining visual grounding, language conditioning, and action prediction within a unified policy model \citep{Octo_2024,OpenVLA_24,PI0,ChatVLA_25}. As these systems scale to broader data mixtures, more tasks, and stronger vision-language backbones, they are increasingly trained on large embodied datasets such as Open X-Embodiment \citep{OXE_24}. These datasets are typically composed of dense robot demonstration trajectories, often collected through teleoperation, where each trajectory records a sequence of observations and actions produced while completing a task. This scaling trend has improved task coverage and generalization, but it also exposes a basic training convention that remains largely unquestioned: dense demonstrations are sampled as if every trajectory frame provided equally useful supervision.

\begin{figure}[!t]
  \centering
  \includegraphics[width=0.40\textwidth]{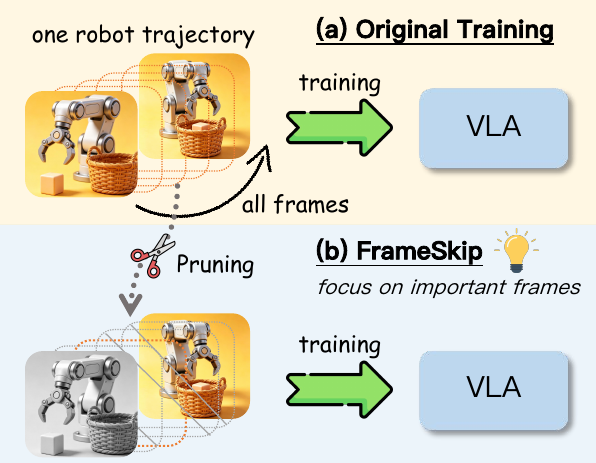}
  \caption{\methodname{} reframes training-time frame pruning as temporal supervision allocation: it reduces exposure to redundant low-change trajectory segments and increases exposure to manipulation-critical transitions.}
    \label{fig:frameskip_intro}
\end{figure}

This convention is mismatched with the temporal structure of robot demonstrations, as illustrated in Figure~\ref{fig:frameskip_intro}. Manipulation trajectories often contain long low-change segments, such as approaching an object, maintaining a grasp, or transporting an object steadily toward a target. In contrast, the moments that define the task outcome are sparse: alignment, contact, grasp closure, release, and abrupt changes in end-effector behavior may occupy only a small fraction of the recorded trajectory. Uniform frame sampling therefore creates a temporal supervision imbalance. Under a fixed optimization budget, rare decision-critical transitions can be diluted by abundant but weakly informative observations.

\begin{figure}[!t]
  \centering
  \includegraphics[width=0.48\textwidth]{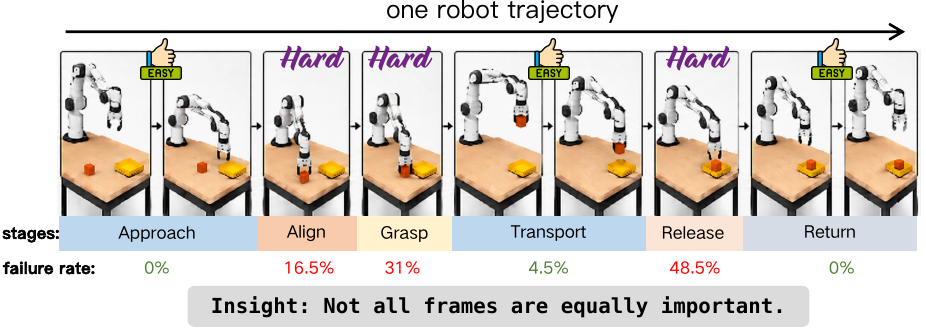}
  \caption{Robot trajectories contain long redundant segments and sparse manipulation-critical transitions, motivating frame selection as a training supervision allocation problem.}
    \label{fig:frameskip_motivation}
\end{figure}

As illustrated in Figure~\ref{fig:frameskip_motivation}, failures are not uniformly distributed along a trajectory: routine stages such as approach and return are often handled reliably, whereas sparse interaction stages such as alignment, grasping, and release exhibit substantially higher failure rates. This stage-wise failure concentration suggests that VLA policies can adapt to dominant smooth motions while remaining brittle at sparse manipulation-critical transitions. We interpret this pattern as global adaptation but local under-supervision, motivating frame selection not as data reduction alone, but as a way to rebalance training toward the moments where policy learning is most fragile.

Existing VLA research has largely addressed scaling through model architecture, action representation, data mixture design, and optimization strategy \citep{OpenVLA_24,OpenVLA-OFT_2025,FAST_25,PI05_25,GR00T_25}. Much less attention has been paid to how supervision is distributed across the frames within each demonstration. Yet this frame-level structure is especially important in embodied data, where trajectories are temporally dense, physically constrained, and dominated by smooth motion. This raises a simple question: can VLA training benefit from reallocating supervision toward the frames that carry the most policy-relevant information?

We therefore view frame selection not merely as a way to reduce data volume, but as a mechanism for reallocating temporal supervision under a fixed optimization budget. In this paper, we present \methodname{}, a data-layer frame selection framework for VLA training. \methodname{} assigns each frame an importance score from lightweight trajectory cues, including action variation, visual-action coherence, task-progress priors, and gripper-transition preservation. It then constructs compressed trajectory views under target retention ratios and remaps training samples toward retained high-importance frames. Importantly, \methodname{} does not modify the VLA architecture, action head, loss function, or inference procedure. This makes \methodname{} a direct way to study frame importance as a training principle rather than as a model-specific architectural change.

We evaluate \methodname{} as a question about the success-retention trade-off of VLA training rather than as a generic frame dropping heuristic. Under matched settings, we compare full-frame training, random frame selection, action-variation-only selection, and progressively stronger importance metrics on RoboCasa-GR1 \citep{RoboCasa_24}, SimplerEnv \citep{SimplerEnv_24}, and LIBERO \citep{libero}. In the main setting, \methodname{} uses a compressed trajectory view that retains 20\% of unique frames and improves the macro-average success rate across the three benchmarks from 66.50\% with full-frame training to 76.15\%, with consistent gains on all three benchmarks.

Our main contributions are as follows:
\begin{itemize}
    \item To our knowledge, we present the first VLA training approach that optimizes supervision at the frame level, identifying temporal supervision imbalance as a practical and underexplored issue in VLA training.
    \item We introduce \methodname{}, an architecture-agnostic data-layer framework that selects more informative training frames using lightweight trajectory cues and gripper-transition preservation.
    \item We provide a systematic empirical study of importance-guided frame retention, including matched-ratio baselines and ablations over retention ratios, importance metrics, and warmup schedules.
\end{itemize}

\section{Related Work}
\label{sec:related}

\textbf{Vision-language-action models.}
VLA models combine visual grounding, language conditioning, and action prediction in a unified policy interface \citep{OpenVLA_24,PI0,ChatVLA_25}. Recent work improves these systems through stronger VLM initialization, action tokenization, diffusion or flow-matching action heads, and large-scale cross-embodiment data \citep{FAST_25,PI05_25,GR00T_25,OXE_24}. These advances generally assume that the training set is consumed at its original temporal density. \methodname{} is complementary: it asks whether the same VLA families can be trained with fewer but more informative frames.

\textbf{Data Curation for Robot Learning.}
Coarse-grained approaches reweight datasets \citep{ReMix_24} or filter trajectories \cite{demonstration-info_25} but treat intra-trajectory frames uniformly. Scizor \cite{SCIZOR_26} curates transitions via a learned task-progress predictor, aiming to remove low-quality and redundant data. \methodname{} differs in objective and mechanism: it does not learn an auxiliary transition-quality model or frame deletion policy, but reallocates training supervision within each trajectory using lightweight cues, including action variation, visual-action coherence, task-progress priors, and gripper-transition preservation, under a controllable retention ratio. TGM-VLA \cite{TGM-VLA_26} addresses keyframe over-sampling in 3D manipulation, but is specific to keyframe-based architectures. \methodname{} operates on raw frames without keyframe structure.

\begin{figure*}[!t]
    \centering
    \includegraphics[width=0.8\textwidth]{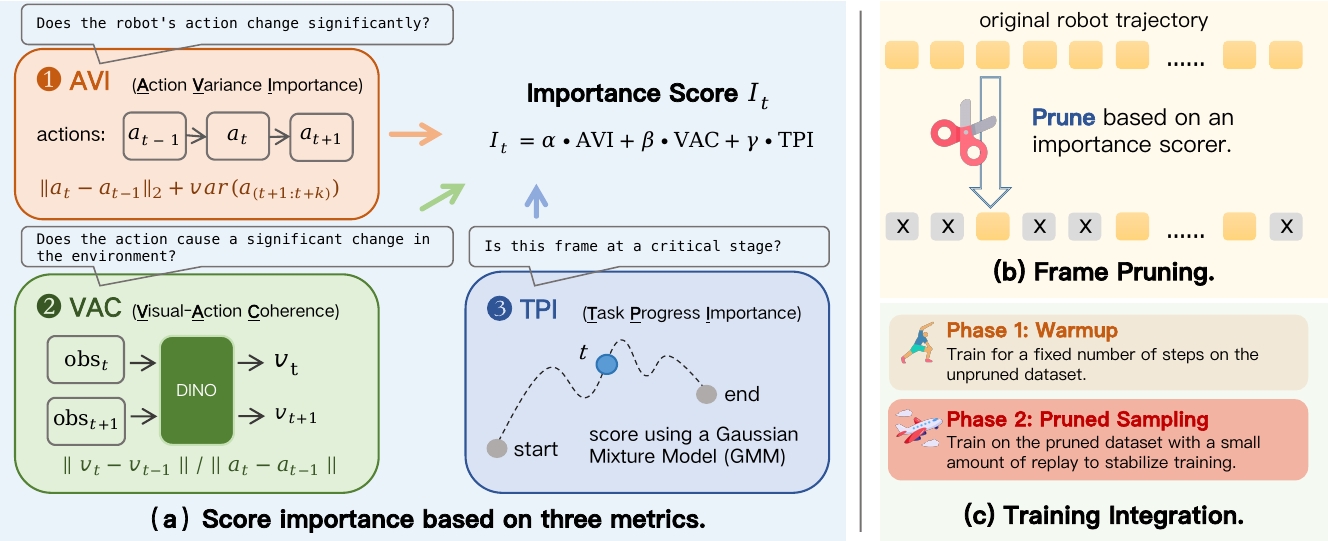}
    \caption{\methodname{} pipeline. \methodname{} scores frames in each demonstration trajectory, retains high-importance frames under a target ratio, and injects the compressed trajectory view into VLA training through dataloader index remapping while leaving the model and inference procedure unchanged.}
    \label{fig:overview}
\end{figure*}

\section{Method}
\label{sec:method}

\subsection{Overview}
\label{sec:method:overview}

\methodname{} is a training-time data-layer framework for reducing temporal redundancy in VLA demonstrations. Given a robot demonstration trajectory, it first computes frame-level importance scores from lightweight trajectory statistics, then precomputes retained frame indices for a set of retention ratios, and finally uses these cached indices to remap dataset queries during training. The VLA model, action head, loss function, and inference procedure are left unchanged. This section formalizes the frame selection problem, describes the importance estimator, presents the ratio-aware pruning rule, and explains how the cached compressed views are integrated into minibatch training.

\subsection{Problem Formulation}
\label{sec:method:problem}

We consider a VLA training set composed of robot demonstration trajectories
$\tau=\{(o_t, a_t, l)\}_{t=1}^{T}$,
where $o_t$ denotes the observation at step $t$, $a_t$ denotes the action, and $l$ denotes the language instruction associated with the trajectory. Standard training uses all frames in $\tau$, implicitly assuming that each timestep contributes equally to learning. \methodname{} challenges this assumption by selecting a subset of frames that is intended to preserve the most informative supervision.

Given a target retention ratio $r \in (0,1]$, our goal is to construct a subset of timestep indices $S_r \subseteq \{1,\dots,T\}$ such that $|S_r| \approx rT$ while preserving the frames that are most useful for learning the policy. The ratio $r$ denotes the fraction of frames retained, rather than the fraction removed. Importantly, \methodname{} is a training-time data transformation: it does not change the VLA model architecture, the action representation, or the inference procedure. Instead, it changes which frames are exposed to the model during training.

\subsection{Frame Importance Estimation}
\label{sec:method:importance}

The core idea of \methodname{} is that trajectory frames should not be treated uniformly. We therefore assign each frame an importance score that combines multiple complementary signals. Intuitively, a frame should receive a higher score if it corresponds to a substantial action change, a visually grounded transition, or a stage of the trajectory where critical interaction is likely to happen. All component scores are min-max normalized within each trajectory before being combined; if a component is constant, it is mapped to a uniform score so that it does not introduce spurious preference.

\paragraph{Action Variation Importance.}
Our first signal captures local action dynamics. Let $a_t$ denote the action at step $t$. We define Action Variation Importance (AVI) as
\begin{equation}
\mathrm{AVI}(t) =
\lVert a_t-a_{t-1}\rVert_2
 + \lambda \,\mathrm{MeanVar}(a_{t+1:t+k}),
\end{equation}
where the first term measures the change relative to the previous action and the second term captures short-range action variation in the next $k$ steps. In our implementation, $k=3$ and $\lambda=0.1$. Near trajectory boundaries, the look-ahead window is truncated to the available timesteps, and the score for the first frame is padded with the first available action-difference value. Frames with large AVI values typically correspond to abrupt motion changes, contact events, grasping, release, or other behavior transitions that are likely to be informative for policy learning.

\paragraph{Visual-Action Coherence.}
Action changes do not always imply meaningful interaction with the environment. To capture visually grounded transitions, \methodname{} incorporates Visual-Action Coherence (VAC):
\begin{equation}
\mathrm{VAC}(t) = \frac{\lVert v_t - v_{t-1} \rVert_2}{\lVert a_t - a_{t-1} \rVert_2 + \epsilon},
\end{equation}
where $v_t$ is a visual feature extracted from observation $o_t$ by a DINOv2 visual encoder. This term gives higher weight to frames where visual change is large relative to the local action change, which is useful for identifying contact or object-motion stages that are not fully captured by action magnitude alone. In all reported \methodname{} experiments, VAC is enabled throughout frame-score preprocessing. To make the offline computation robust and affordable, we compute VAC on sparsely sampled video frames, interpolate the resulting scores back to the action sequence length, and clip extreme VAC values at the 95th percentile before normalization.

\paragraph{Task Progress Importance.}
Some interaction events are sparse but tend to occur in characteristic regions of a task trajectory. To encode this weak structural prior, we define Task Progress Importance (TPI) over the normalized progress
$p_t=(t-1)/(T-1)$.
In the main experiments, we use a dataset-adaptive progress prior. Specifically, for each benchmark, we fit a one-dimensional Gaussian mixture model (GMM) to the normalized progress locations of manipulation-critical stage centers annotated from a small subset of training trajectories:
\begin{equation}
q(p) = \sum_{m=1}^{M} \pi_m \, \mathcal{N}(p;\mu_m,\sigma_m^2),
\end{equation}
and define
\begin{equation}
\mathrm{TPI}(t) =
\frac{q(p_t)}{\max_{s \in \{1,\dots,T\}} q(p_s)}.
\end{equation}
This dataset-adaptive prior captures task-specific stage structure while keeping frame scoring independent of the VLA model and policy objective. The stage annotations are used only to estimate the offline progress prior during preprocessing and are not provided to the policy during training or evaluation.

When such annotations are unavailable, \methodname{} can use a simpler dataset-agnostic Gaussian prior:
\begin{equation}
\mathrm{TPI}(t) =
\exp \left( -\frac{(p_t-0.5)^2}{\sigma^2} \right),
\quad
p_t=\frac{t-1}{T-1}.
\end{equation}
This fallback assumes that manipulation-critical stages are more likely to occur near the middle of a trajectory and requires no stage annotations; we use $\sigma^2=0.2$ for this Gaussian variant.

\paragraph{Combined score and gripper-transition preservation.}
We combine the signals into a single frame score:
\begin{equation}
I(t) = \alpha \, \widehat{\mathrm{AVI}}(t) + \beta \, \widehat{\mathrm{VAC}}(t) + \gamma \, \widehat{\mathrm{TPI}}(t),
\label{eq:importance}
\end{equation}
where $\widehat{\cdot}$ denotes min-max normalized scores and $\alpha,\beta,\gamma$ are scalar weights. In our default setting, AVI provides the dominant signal, while VAC and TPI act as auxiliary cues; we use $\alpha=0.6$, $\beta=0.2$, and $\gamma=0.2$ unless otherwise specified. Ablation variants may remove VAC to isolate its contribution, but the full \methodname{} configuration used in the main experiments enables VAC.

For manipulation tasks, some of the most important moments coincide with gripper or end-effector state transitions. The gripper-aware variant therefore multiplies the combined score by a factor determined by the absolute change in the gripper or end-effector state dimensions specified by each benchmark action schema. When such dimensions are unavailable, this factor falls back to the action-variation signal already captured by AVI. This design does not introduce a new model component; it simply injects a task-relevant event prior into the scoring function so that contact-related stages are less likely to be removed during pruning.

\subsection{Ratio-Aware Frame Pruning}
\label{sec:method:pruning}

Once importance scores are computed, \methodname{} prunes frames according to a target retention ratio $r$. For a trajectory of length $T$, the target number of retained frames is
\begin{equation}
K_r=\max(K_{\min}, \lfloor rT \rfloor),
\end{equation}
where $K_{\min}$ prevents very short compressed trajectories. We first compute a threshold based on the empirical $(1-r)$-quantile of the importance scores and retain frames whose score exceeds that threshold:
\begin{equation}
S_r = \{t \mid I(t) \geq \theta_r\},
\end{equation}
where $\theta_r=\mathrm{Quantile}(I,1-r)$, so the candidate set approximately contains the top $rT$ frames.

The pruning procedure additionally enforces several practical constraints. First, when gripper-transition preservation is enabled, the pruner explicitly retains the first frame, the last frame, gripper or end-effector transition frames, and frames whose action changes fall in the top decile of the trajectory. Second, if the quantile rule keeps too many or too few frames relative to $K_r$, the pruner selects or adds frames by descending importance until the target count is met. Third, we optionally apply a temporal consistency constraint that fills unusually large gaps between consecutive retained frames. This avoids pathological cases in which a trajectory becomes too temporally discontinuous after pruning, at the cost of a slightly higher actual retention ratio.

In practice, \methodname{} supports multiple retention ratios for the same trajectory. We therefore precompute and cache pruning results for a configured superset of ratios. Each trajectory cache stores the retained indices and the actual achieved ratio for each configured setting, allowing the training pipeline to switch between compressed views without recomputing frame scores. The cache is keyed by the importance and pruning configuration; a separate list of training ratios can be chosen as a subset of the cached ratios to reuse the same cache across multiple schedules.

\subsection{Sampling Strategy}
\label{sec:method:sampling}

\methodname{} uses compressed trajectories as the main source of supervision after an initial full-frame warmup. The motivation is to make the policy learn primarily from high-importance frames, while still preserving occasional access to the original temporal density. This gives the training process two complementary signals: compressed mini-batches emphasize decision-relevant moments, whereas full-frame mini-batches act as an anchor that refreshes the broader trajectory context and reduces the risk of overfitting to overly sparse transitions.

\textbf{Warmup.}
During the first $N_{\mathrm{warm}}$ optimization steps, \methodname{} uses the identity view with $r=1.0$, which is equivalent to standard full-frame training. This stage gives the policy a stable initialization from dense temporal supervision before the frame-pruned views are introduced.

\textbf{Pruned Sampling with Full-Frame Anchors.}
After warmup, most mini-batches are drawn from a frame-pruned view with a target retention ratio $r<1.0$, so the effective training distribution is biased toward frames selected by the importance estimator. A small fraction of mini-batches are instead drawn from the full-frame view $r=1.0$. We use this mixture to preserve global trajectory coverage while still concentrating supervision on high-importance frames. Under a fixed number of optimization steps, this schedule changes which timesteps dominate the gradient signal rather than changing the policy objective. In our main setting, \methodname{} uses a compressed view with $r=0.2$, retaining 20\% of unique frames from each trajectory and pruning the remaining 80\% within that view. For every five pruned mini-batches, we insert one full-frame mini-batch as a context anchor. This schedule treats full-frame samples not as the default training signal, but as periodic context refreshes that stabilize learning under aggressive temporal compression.

\subsection{Training Integration}
\label{sec:method:integration}

\methodname{} is designed as a data-layer intervention. Rather than rewriting the original dataset or modifying the VLA model, we keep the original trajectory index space unchanged and perform frame selection through index remapping at data loading time.

Concretely, each sampled training step is first mapped to its trajectory and original timestep through the standard LeRobot dataset index. Given the active retention ratio, the dataloader retrieves the cached retained indices for that trajectory and uses binary search to map the requested timestep to the first retained timestep that is not earlier than the request, falling back to the final retained timestep at the end of the trajectory. The resulting frame is then loaded with the original data access function and passed through the standard transform and collation pipeline. The returned sample also records the active ratio, the original timestep, and the remapped timestep for logging and analysis.

This design has two practical benefits. First, \methodname{} is architecture-agnostic: the same mechanism can be used with different VLA backbones and action heads. Second, it preserves compatibility with existing dataset mixtures and sampling weights, because the apparent dataset length and trajectory index space remain unchanged. Changing the active retention ratio only changes the dataset index mapping rather than the optimization objective or the surrounding trainer logic.

\section{Experiments}
\label{sec:experiments}

\subsection{Experimental Setup}
\label{sec:experimental_setup}

\paragraph{Models and Framework.}
We instantiate all VLA policies in the StarVLA framework \citep{starvla_2025} with a two-expert architecture. The understanding expert is initialized from Qwen3-4B-VL-Instruct \citep{Qwen3-VL}, which encodes the language instruction and visual observation into multimodal hidden states. The action expert is a randomly initialized Diffusion Transformer (DiT) \citep{DiT_23} that generates continuous robot actions with a flow-matching objective. Concretely, the last hidden states of the VLM are passed as conditioning features to the action expert, allowing the policy to preserve the semantic and visual grounding ability of the VLM while learning benchmark-specific action generation from robot demonstrations.

\paragraph{Training Details.}
For each benchmark, we train the VLA policy on the corresponding benchmark-specific training set for a fixed number of optimization steps. The number of training steps is adjusted according to the size of each benchmark dataset, while the global batch size is kept fixed at 128 across all runs. All experiments are conducted on 8 NVIDIA H100 GPUs with DeepSpeed ZeRO-2 distributed training \citep{deepspeed_2020}. Unless otherwise specified, \methodname{} uses a retention ratio of $r=0.2$ and a 5:1 schedule between pruned mini-batches and full-frame anchor mini-batches. The same model architecture, optimizer setting, and remaining training configuration are used across compared methods so that differences can be attributed to the frame selection strategy rather than to changes in the underlying VLA training recipe. To facilitate reproducibility and future work on frame-level VLA training data optimization, we will publicly release the training code, frame-selection pipeline, and model checkpoints. Additional implementation and hyperparameter details are provided in Appendix~\ref{app:implementation}.

\paragraph{Benchmarks.}
We evaluate \methodname{} on three simulation benchmarks: RoboCasa-GR1, SimplerEnv \citep{SimplerEnv_24}, and LIBERO \citep{libero}. These benchmarks cover different robot embodiments, manipulation settings, and evaluation protocols. Since embodied benchmarks are tied to different robot morphologies, controllers, observation spaces, and action conventions, each benchmark requires VLA training on robot data from the corresponding embodiment. This setting tests whether \methodname{} can be applied as a data-layer frame pruning method across multiple embodied data regimes rather than only within a single robot platform.

\subsection{Simulation Benchmarks}
\label{sec:simulation_benchmarks}

\textbf{RoboCasa-GR1.}
RoboCasa-GR1 is a tabletop manipulation benchmark built on RoboCasa \citep{RoboCasa_24}, where a GR1 robot performs bimanual manipulation with two dexterous hands. We evaluate on 24 tabletop tasks and train with the 24K GR1 teleoperation simulation demonstrations released by NVIDIA. This benchmark tests multi-task VLA learning and dexterous-hand control. The main results are shown in Table~\ref{tab:robocasa_gr1_main}, and the full 24-task results are provided in Appendix~\ref{app:robocasa_full}.

\begin{table}[!t]
    \centering
    \small
    \setlength{\tabcolsep}{3.5pt}
    \renewcommand{\arraystretch}{1.15}
    \caption{
    RoboCasa-GR1 simulation results on four representative pick-and-place tasks.
    The omitted columns indicate additional RoboCasa-GR1 tasks; Avg. is computed over all 24 tasks rather than only the shown tasks.
    The first block lists representative VLA systems, while the final block isolates the controlled comparison between full-frame training and \methodname{}.
    }
    \resizebox{\linewidth}{!}{
    \rowcolors{2}{gray!15}{white}
    \begin{tabular}{lcccccc}
        \toprule
        \textbf{Method} & \textbf{PnP Bottle} & \textbf{PnP Can} & \textbf{PnP Cup} & \textbf{PnP Milk} & $\cdots$ & \textbf{Avg.} \\
        \midrule
        GR00T N1.5~\citep{GR00T_25}        & 54.0 & 50.0 & 38.0 & 60.0 & $\cdots$ & 48.2 \\
        GR00T N1.6~\citep{GR00T_N1.6}        & 51.5 & 13.0 &  8.5 & 14.0 & $\cdots$ & 47.6 \\
        TwinBrainVLA~\citep{TwinBrainVLA}      & 74.0 & 72.0 & 52.0 & 60.0 & $\cdots$ & 54.6 \\
        PhysBrain~\citep{PhysBrain_25}         & 74.0 & 68.0 & 42.0 & 54.0 & $\cdots$ & 50.0 \\
        LangForce~\citep{LangForce}         & 72.0 & 78.0 & 46.0 & 56.0 & $\cdots$ & 52.6 \\
        ABot-M0~\citep{ABot-M0_26}           & 86.0 & 74.0 & 48.0 & 46.0 & $\cdots$ & \underline{58.3} \\
        \midrule
        Full-Frame Training & 46.0 & 80.0 & 54.0 & 48.0 & $\cdots$ & 47.8 \\
        \textbf{\methodname{} (ours)}       & 74.0 & 80.0 & 46.0 & 60.0 & $\cdots$ & \textbf{59.5} \\
        \bottomrule
    \end{tabular}}
    \label{tab:robocasa_gr1_main}
\end{table}

\textbf{SimplerEnv.}
SimplerEnv evaluates WidowX manipulation policies in simulation \citep{SimplerEnv_24}. We use four evaluation tasks whose scenes and instructions are held out from training, making the benchmark a test of out-of-domain generalization. Following the standard setting, we train on the BridgeV2 real-robot dataset and evaluate in SimplerEnv simulation. The results are shown in Table~\ref{tab:simplerenv_main}.

\begin{table}[!t]
    \centering
    \small
    \setlength{\tabcolsep}{3.0pt}
    \renewcommand{\arraystretch}{1.15}
    \caption{
    SimplerEnv simulation results on four held-out WidowX manipulation tasks.
    We report success rates (\%) for each task and the average across the four tasks.
    The first block lists representative VLA systems, while the final block isolates the controlled comparison between full-frame training and \methodname{}.
    }
    \resizebox{\linewidth}{!}{
    \rowcolors{2}{gray!15}{white}
    \begin{tabular}{lccccc}
        \toprule
        \textbf{Method} &
        \makecell{\textbf{Put Spoon}\\\textbf{on Towel}} &
        \makecell{\textbf{Put Carrot}\\\textbf{on Plate}} &
        \makecell{\textbf{Stack Green Block}\\\textbf{on Yellow Block}} &
        \makecell{\textbf{Put Eggplant}\\\textbf{in Yellow Basket}} &
        \textbf{Avg.} \\
        \midrule
        OpenVLA~\citep{OpenVLA_24}     & 4.2  & 0.0  & 0.0 & 12.5 & 4.2  \\
        RoboVLM~\citep{RoboVLM_2024}     & 50.0 & 37.5 & 0.0 & 83.3 & 42.7 \\
        ThinkAct~\citep{ThinkAct_25}    & 58.3 & 37.5 & 8.7 & 70.8 & 43.8 \\
        SpatialVLA~\citep{SpatialVLA_2025}  & 20.8 & 20.8 & 25.0 & 70.8 & 34.4 \\
        CogACT~\citep{CogACT_2024}      & 71.7 & 50.8 & 15.0 & 67.5 & 51.3 \\
        VideoVLA~\citep{VideoVLA_2025}    & 75.0 & 20.8 & 45.8 & 70.8 & 53.1 \\
        $\pi_0$~\citep{PI0}     & 29.1 & 0.0  & 16.6 & 62.5 & 27.1 \\
        $\pi_{0.5}$~\citep{PI05_25} & 49.3 & 64.7 & 44.7 & 69.7 & 57.1 \\
        GR00T N1.6~\citep{GR00T_N1.6}  & 64.5 & 65.5 & 5.5  & 93.0 & 57.1 \\
        VLA-JEPA~\citep{VLA-JEPA_26}    & 75.0 & 70.8 & 12.5 & 70.8 & 57.3 \\
        TwinBrainVLA~\citep{TwinBrainVLA} & 87.5 & 58.3 & 33.3 & 79.1 & 64.5 \\
        LangForce~\citep{LangForce}    & 89.6 & 63.8 & 33.3 & 79.2 & \underline{66.5} \\
        \midrule
        Full-Frame Training       & 87.5 & 50.0 & 29.2 & 54.2 & 55.2 \\
        \textbf{\methodname{} (ours)} & 90.63 & 54.17 & 45.59 & 95.83 & \textbf{71.55} \\
        \bottomrule
    \end{tabular}}
    \label{tab:simplerenv_main}
\end{table}

\textbf{LIBERO.}
LIBERO is a Franka-based simulation benchmark for language-conditioned manipulation \citep{libero}. We evaluate on four task suites and train with the official expert demonstrations provided by the benchmark. LIBERO complements RoboCasa-GR1 and SimplerEnv by testing \methodname{} on a standardized single-arm embodiment with expert trajectories. The results are shown in Table~\ref{tab:libero_main}.

\begin{table}[!t]
    \centering
    \small
    \setlength{\tabcolsep}{4.0pt}
    \renewcommand{\arraystretch}{1.15}
    \caption{
    LIBERO simulation results on four task suites.
    We report success rates (\%) on Spatial, Object, Goal, and Long, together with the average across the four suites.
    The first block lists representative policy/VLA systems, while the final block isolates the controlled comparison between full-frame training and \methodname{}.
    }
    \resizebox{\linewidth}{!}{
    \rowcolors{2}{gray!15}{white}
    \begin{tabular}{lccccc}
        \toprule
        \textbf{Method} & \textbf{L-Spatial} & \textbf{L-Object} & \textbf{L-Goal} & \textbf{L-Long} & \textbf{Avg.} \\
        \midrule
        Diffusion Policy~\citep{DiffusionPolicy_23} & 78.5 & 87.5 & 73.5 & 64.8 & 76.1 \\
        OpenVLA~\citep{OpenVLA_24}          & 84.7 & 88.4 & 79.2 & 53.7 & 76.5 \\
        SpatialVLA~\citep{SpatialVLA_2025}       & 88.2 & 89.9 & 78.6 & 55.5 & 78.1 \\
        CoT-VLA~\citep{CoT-VLA}          & 87.5 & 91.6 & 87.6 & 69.0 & 83.9 \\
        GR00T N1~\citep{GR00T_25}         & 94.4 & 97.6 & 93.0 & 90.6 & 93.9 \\
        F1~\citep{F1_25}               & 98.2 & 97.8 & 95.4 & 91.3 & 95.7 \\
        InternVLA-M1~\citep{InternVLA_M1_25}     & 98.0 & 99.0 & 93.8 & 92.6 & 95.9 \\
        $\pi_0$~\citep{PI0}          & 98.0 & 96.8 & 94.4 & 88.4 & 94.4 \\
        $\pi_{0.5}$~\citep{PI05_25}      & 98.8 & 98.2 & 98.0 & 92.4 & 96.9 \\
        GR00T N1.6~\citep{GR00T_N1.6}       & 97.7 & 98.5 & 97.5 & 94.4 & \underline{97.0} \\
        \midrule
        Full-Frame Training       & 97.8 & 98.8 & 97.4 & 92.0 & 96.5 \\
        \textbf{\methodname{} (ours)} & 98.6 & 99.0 & 98.2 & 93.8 & \textbf{97.4} \\
        \bottomrule
    \end{tabular}}
    \label{tab:libero_main}
\end{table}

\textbf{Results.}
Across the three simulation benchmarks, \methodname{} consistently improves over full-frame training under the same VLA architecture and training recipe. It improves the macro-average success rate across RoboCasa-GR1, SimplerEnv, and LIBERO from 66.50\% to 76.15\% while using a compressed trajectory view that retains 20\% of unique frames in the main setting, suggesting that reallocating supervision toward informative frames is a useful training signal rather than merely a data reduction heuristic.

\subsection{Ablation Studies}
\label{sec:ablation}

We conduct ablation studies to isolate the design choices behind \methodname{} and to test whether its gains come from principled frame selection rather than from using fewer training frames alone. The ablations are organized around three questions: how much temporal supervision should be retained, which importance cues are responsible for selecting useful frames, and how much dense full-frame training is needed before introducing compressed trajectory views.

\textbf{Effect of retention ratio.}
The retention ratio controls the central trade-off in \methodname{}: retaining more frames preserves denser trajectory context, while retaining fewer frames increases the concentration of supervision on high-importance moments. On RoboCasa-GR1, we evaluate retention ratios $r \in \{10\%,20\%,30\%,40\%,50\%,60\%,100\%\}$ with the same model and training budget, using $r=100\%$ as the full-frame reference. This ablation tests whether performance peaks at a moderate compression level and whether aggressive pruning removes context needed for stable policy learning. As shown in Table~\ref{tab:ablation_retention}, all pruned settings outperform full-frame training, with the best result at $r=50\%$ and strong performance already at $r=20\%$--$30\%$, supporting our central claim that reallocating supervision toward informative frames can be more effective than exposing the model to every temporally redundant frame.

\begin{table}[!t]
    \centering
    \small
    \setlength{\tabcolsep}{4.0pt}
    \renewcommand{\arraystretch}{1.15}
    \caption{
    Ablation on the retention ratio $r$.
    We report the average success rate (\%) across the 24 RoboCasa-GR1 tasks.
    }
    \resizebox{\linewidth}{!}{
    \rowcolors{2}{gray!15}{white}
    \begin{tabular}{lccccccc}
        \toprule
        \textbf{Retention $r$} & \textbf{10\%} & \textbf{20\%} & \textbf{30\%} & \textbf{40\%} & \textbf{50\%} & \textbf{60\%} & \textbf{100\%} \\
        \midrule
        \textbf{RoboCasa-GR1 Avg.} & 55.00 & 59.50 & 59.50 & 56.75 & \textbf{59.75} & 55.92 & 47.80 \\
        \bottomrule
    \end{tabular}}
    \label{tab:ablation_retention}
\end{table}

\textbf{Effect of importance metric.}
To understand which scoring cues matter, we compare several frame selection variants under the same retention ratio on RoboCasa-GR1, SimplerEnv, and LIBERO. The random variant retains frames without using trajectory information and serves as a pruning-only control. The AVI-only variant uses action variation as the sole importance signal. We then add task-progress information (AVI+TPI), visual-action coherence (AVI+VAC), and their combination (AVI+VAC+TPI). Finally, \methodname{} Full uses the complete scoring and preservation strategy, including gripper-transition preservation. This ablation tests whether each cue contributes complementary information and whether the full method outperforms simpler action-only or randomly pruned views. The gains over random pruning and action-only variants indicate that the benefit comes from where supervision is allocated, not simply from seeing fewer frames. The results are reported in Table~\ref{tab:ablation_importance}.

\begin{table}[!t]
    \centering
    \small
    \setlength{\tabcolsep}{3.5pt}
    \renewcommand{\arraystretch}{1.15}
    \caption{
    Ablation on the frame importance metric.
    All variants use the same retention ratio and training schedule; only the frame scoring rule is changed.
    We report success rates (\%) on RoboCasa-GR1, SimplerEnv, and LIBERO, together with their average.
    }
    \resizebox{\linewidth}{!}{
    \rowcolors{2}{gray!15}{white}
    \begin{tabular}{lcccc}
        \toprule
        \textbf{Metric Variant} & \textbf{RoboCasa-GR1} & \textbf{SimplerEnv} & \textbf{LIBERO} & \textbf{Avg.} \\
        \midrule
        Random & 47.67 & 56.51 & 96.3 & 66.83 \\
        AVI & 54.25 & 57.29 & 97.05 & 69.53 \\
        AVI+TPI & 57.42 & 59.90 & 97.00 & 71.44 \\
        AVI+VAC & 58.75 & 65.08 & 97.15 & 73.66 \\
        AVI+VAC+TPI & 59.00 & 67.33 & 97.2 & 74.51 \\
        \textbf{\methodname{} Full} & 59.50 & 71.55 & 97.4 & 76.15 \\
        \bottomrule
    \end{tabular}}
    \label{tab:ablation_importance}
\end{table}

\textbf{Effect of warmup steps.}
We also study the sensitivity of \methodname{} to the length of the initial full-frame warmup on RoboCasa-GR1. As shown in Table~\ref{tab:ablation_warmup}, changing the warmup length from 2500 to 15000 optimization steps has only a modest effect on the final average success rate, suggesting that \methodname{} is not highly sensitive to this hyperparameter. The best result is obtained with 5000 warmup steps. This indicates that a short but sufficient full-frame warmup can establish basic visual-action grounding, after which the remaining training can focus more heavily on pruned frames selected by \methodname{}.

\begin{table}[!t]
    \centering
    \small
    \setlength{\tabcolsep}{4.0pt}
    \renewcommand{\arraystretch}{1.15}
    \caption{
    Ablation on the number of full-frame warmup steps.
    After warmup, all variants use the same retention ratio and pruned/full-frame mini-batch schedule.
    We report the average success rate (\%) across the 24 RoboCasa-GR1 tasks.
    }
    \resizebox{\linewidth}{!}{
    \rowcolors{2}{gray!15}{white}
    \begin{tabular}{lcccccc}
        \toprule
        \textbf{Warmup Steps} & \textbf{2500} & \textbf{5000} & \textbf{7500} & \textbf{10000} & \textbf{12500} & \textbf{15000} \\
        \midrule
        \textbf{RoboCasa-GR1 Avg.} & 58.42 & \textbf{59.50} & 59.08 & 58.75 & 58.33 & 58.25 \\
        \bottomrule
    \end{tabular}}
    \label{tab:ablation_warmup}
\end{table}

\section{Conclusion}
\label{sec:conclusion}

We presented \methodname{}, a training-time frame pruning framework for VLA models. The method is motivated by a simple observation: robot trajectories contain structured temporal redundancy, and not every frame contributes equally to policy learning. By combining action variation, visual-action coherence, task-progress priors, and gripper-transition preservation, \methodname{} selects more informative frames under a target retention budget while leaving the VLA architecture unchanged. Across RoboCasa-GR1, SimplerEnv, and LIBERO, \methodname{} improves the macro-average success rate across the three benchmarks from 66.50\% to 76.15\% while using a compressed trajectory view that retains 20\% of unique frames in the main setting, showing that frame-level supervision allocation can be a practical lever for VLA training. The broader goal is to make frame importance a first-class object in embodied multimodal learning.

\bibliography{custom}

@String(CoRL = {Annual Conference on Robot Learning (CoRL)})

@String(ICCV = {Proceedings of the IEEE/CVF International Conference on Computer Vision (ICCV)})

@String(ICRA = {IEEE International Conference on Robotics and Automation (ICRA)})

@String(EMNLP = {Proceedings of the Empirical Methods in Natural Language Processing (EMNLP)})

@inproceedings{OpenVLA_24,
  title={OpenVLA: An Open-Source Vision-Language-Action Model},
  author={Moo Jin Kim and Karl Pertsch and Siddharth Karamcheti and Ted Xiao and Ashwin Balakrishna and Suraj Nair and Rafael Rafailov and Ethan P Foster and Pannag R Sanketi and Quan Vuong and Thomas Kollar and Benjamin Burchfiel and Russ Tedrake and Dorsa Sadigh and Sergey Levine and Percy Liang and Chelsea Finn},
  booktitle=CoRL,
  year={2024}
}

@inproceedings{DiffusionPolicy_23,
	title={Diffusion Policy: Visuomotor Policy Learning via Action Diffusion},
	author={Chi, Cheng and Feng, Siyuan and Du, Yilun and Xu, Zhenjia and Cousineau, Eric and Burchfiel, Benjamin and Song, Shuran},
	booktitle={Proceedings of Robotics: Science and Systems (RSS)},
	year={2023}
}

@article{PI0,
  title={{$\pi_0$}: A Vision-Language-Action Flow Model for General Robot Control},
  author={Black, Kevin and Brown, Noah and Driess, Danny and Esmail, Adnan and Equi, Michael and Finn, Chelsea and Fusai, Niccolo and Groom, Lachy and Hausman, Karol and Ichter, Brian and others},
  journal={arXiv preprint arXiv:2410.24164},
  year={2024}
}

@misc{PI05_25,
  title={$\pi_{0.5}$: a Vision-Language-Action Model with Open-World Generalization}, 
  author={Physical Intelligence and Kevin Black and Noah Brown and James Darpinian and Karan Dhabalia and Danny Driess and Adnan Esmail and Michael Equi and Chelsea Finn and Niccolo Fusai and Manuel Y. Galliker and Dibya Ghosh and Lachy Groom and Karol Hausman and Brian Ichter and Szymon Jakubczak and Tim Jones and Liyiming Ke and Devin LeBlanc and Sergey Levine and Adrian Li-Bell and Mohith Mothukuri and Suraj Nair and Karl Pertsch and Allen Z. Ren and Lucy Xiaoyang Shi and Laura Smith and Jost Tobias Springenberg and Kyle Stachowicz and James Tanner and Quan Vuong and Homer Walke and Anna Walling and Haohuan Wang and Lili Yu and Ury Zhilinsky},
  year={2025},
  eprint={2504.16054},
  archivePrefix={arXiv},
  primaryClass={cs.LG}
}

@misc{GR00T_25,
      title={GR00T N1: An Open Foundation Model for Generalist Humanoid Robots}, 
      author={NVIDIA and : and Johan Bjorck and Fernando Castañeda and Nikita Cherniadev and Xingye Da and Runyu Ding and Linxi "Jim" Fan and Yu Fang and Dieter Fox and Fengyuan Hu and Spencer Huang and Joel Jang and Zhenyu Jiang and Jan Kautz and Kaushil Kundalia and Lawrence Lao and Zhiqi Li and Zongyu Lin and Kevin Lin and Guilin Liu and Edith Llontop and Loic Magne and Ajay Mandlekar and Avnish Narayan and Soroush Nasiriany and Scott Reed and You Liang Tan and Guanzhi Wang and Zu Wang and Jing Wang and Qi Wang and Jiannan Xiang and Yuqi Xie and Yinzhen Xu and Zhenjia Xu and Seonghyeon Ye and Zhiding Yu and Ao Zhang and Hao Zhang and Yizhou Zhao and Ruijie Zheng and Yuke Zhu},
      year={2025},
      eprint={2503.14734},
      archivePrefix={arXiv},
      primaryClass={cs.RO}
}

@misc{GR00T_25_arxiv,
      title={GR00T N1: An Open Foundation Model for Generalist Humanoid Robots}, 
      author={NVIDIA and : and Johan Bjorck and Fernando Castañeda and Nikita Cherniadev and Xingye Da and Runyu Ding and Linxi "Jim" Fan and Yu Fang and Dieter Fox and Fengyuan Hu and Spencer Huang and Joel Jang and Zhenyu Jiang and Jan Kautz and Kaushil Kundalia and Lawrence Lao and Zhiqi Li and Zongyu Lin and Kevin Lin and Guilin Liu and Edith Llontop and Loic Magne and Ajay Mandlekar and Avnish Narayan and Soroush Nasiriany and Scott Reed and You Liang Tan and Guanzhi Wang and Zu Wang and Jing Wang and Qi Wang and Jiannan Xiang and Yuqi Xie and Yinzhen Xu and Zhenjia Xu and Seonghyeon Ye and Zhiding Yu and Ao Zhang and Hao Zhang and Yizhou Zhao and Ruijie Zheng and Yuke Zhu},
      year={2025},
      eprint={2503.14734},
      archivePrefix={arXiv},
      primaryClass={cs.RO}
}

@inproceedings{ChatVLA_25,
    title={ChatVLA: Unified Multimodal Understanding and Robot Control with Vision-Language-Action Model},
    author={Zhongyi Zhou and Yichen Zhu and Minjie Zhu and Junjie Wen and Ning Liu and Zhiyuan Xu and Weibin Meng and Ran Cheng and Yaxin Peng and Chaomin Shen and Feifei Feng},
    booktitle=EMNLP,
    year="2025",
    publisher = "Association for Computational Linguistics"
}

@InProceedings{DiT_23,
    author    = {Peebles, William and Xie, Saining},
    title     = {Scalable Diffusion Models with Transformers},
    booktitle = ICCV,
    month     = {October},
    year      = {2023},
    pages     = {4195-4205}
}

@misc{Qwen3-VL,
  title={Qwen3-VL Technical Report}, 
  author={Shuai Bai and Yuxuan Cai and Ruizhe Chen and Keqin Chen and Xionghui Chen and Zesen Cheng and Lianghao Deng and Wei Ding and Chang Gao and Chunjiang Ge and Wenbin Ge and Zhifang Guo and Qidong Huang and Jie Huang and Fei Huang and Binyuan Hui and Shutong Jiang and Zhaohai Li and Mingsheng Li and Mei Li and Kaixin Li and Zicheng Lin and Junyang Lin and Xuejing Liu and Jiawei Liu and Chenglong Liu and Yang Liu and Dayiheng Liu and Shixuan Liu and Dunjie Lu and Ruilin Luo and Chenxu Lv and Rui Men and Lingchen Meng and Xuancheng Ren and Xingzhang Ren and Sibo Song and Yuchong Sun and Jun Tang and Jianhong Tu and Jianqiang Wan and Peng Wang and Pengfei Wang and Qiuyue Wang and Yuxuan Wang and Tianbao Xie and Yiheng Xu and Haiyang Xu and Jin Xu and Zhibo Yang and Mingkun Yang and Jianxin Yang and An Yang and Bowen Yu and Fei Zhang and Hang Zhang and Xi Zhang and Bo Zheng and Humen Zhong and Jingren Zhou and Fan Zhou and Jing Zhou and Yuanzhi Zhu and Ke Zhu},
  journal={arXiv preprint arXiv:2511.21631},
  year={2025}
}

@inproceedings{SimplerEnv_24,
  title={Evaluating Real-World Robot Manipulation Policies in Simulation},
  author={Xuanlin Li and Kyle Hsu and Jiayuan Gu and Oier Mees and Karl Pertsch and Homer Rich Walke and Chuyuan Fu and Ishikaa Lunawat and Isabel Sieh and Sean Kirmani and Sergey Levine and Jiajun Wu and Chelsea Finn and Hao Su and Quan Vuong and Ted Xiao},
  booktitle=CoRL,
  year={2024},
}

@inproceedings{RoboCasa_24,
  title={RoboCasa: Large-Scale Simulation of Everyday Tasks for Generalist Robots},
  author={Soroush Nasiriany and Abhiram Maddukuri and Lance Zhang and Adeet Parikh and Aaron Lo and Abhishek Joshi and Ajay Mandlekar and Yuke Zhu},
  booktitle={Robotics: Science and Systems},
  year={2024}
}

@misc{starvla_2025,
  title        = {StarVLA: A Lego-like Codebase for Vision-Language-Action Model Developing},
  author       = {starVLA},
  year         = {2025},
  month        = {1},
  version      = {1.2.0},
  url          = {https://github.com/starVLA/starVLA},
  doi          = {10.5281/zenodo.18264214},
  howpublished = {GitHub repository},
  publisher    = {GitHub}
}

@inproceedings{OXE_24,
  title={Open x-embodiment: Robotic learning datasets and rt-x models: Open x-embodiment collaboration},
  author={O’Neill, Abby and Rehman, Abdul and Maddukuri, Abhiram and Gupta, Abhishek and Padalkar, Abhishek and Lee, Abraham and Pooley, Acorn and Gupta, Agrim and Mandlekar, Ajay and Jain, Ajinkya and others},
  booktitle={2024 IEEE International Conference on Robotics and Automation (ICRA)},
  pages={6892--6903},
  year={2024},
  organization={IEEE}
}

@misc{Octo_2024,
  title={Octo: An Open-Source Generalist Robot Policy}, 
  author={Octo Model Team and Dibya Ghosh and Homer Walke and Karl Pertsch and Kevin Black and Oier Mees and Sudeep Dasari and Joey Hejna and Tobias Kreiman and Charles Xu and Jianlan Luo and You Liang Tan and Lawrence Yunliang Chen and Pannag Sanketi and Quan Vuong and Ted Xiao and Dorsa Sadigh and Chelsea Finn and Sergey Levine},
  year={2024},
  eprint={2405.12213},
  archivePrefix={arXiv},
  primaryClass={cs.RO},
}

@misc{OpenVLA-OFT_2025,
  title={Fine-Tuning Vision-Language-Action Models: Optimizing Speed and Success}, 
  author={Moo Jin Kim and Chelsea Finn and Percy Liang},
  year={2025},
  eprint={2502.19645},
  archivePrefix={arXiv},
  primaryClass={cs.RO},
}

@misc{RoboVLM_2024,
      title={Towards Generalist Robot Policies: What Matters in Building Vision-Language-Action Models}, 
      author={Xinghang Li and Peiyan Li and Minghuan Liu and Dong Wang and Jirong Liu and Bingyi Kang and Xiao Ma and Tao Kong and Hanbo Zhang and Huaping Liu},
      year={2024},
      eprint={2412.14058},
      archivePrefix={arXiv},
      primaryClass={cs.RO},
}

@misc{SpatialVLA_2025,
      title={SpatialVLA: Exploring Spatial Representations for Visual-Language-Action Model}, 
      author={Delin Qu and Haoming Song and Qizhi Chen and Yuanqi Yao and Xinyi Ye and Yan Ding and Zhigang Wang and JiaYuan Gu and Bin Zhao and Dong Wang and Xuelong Li},
      year={2025},
      eprint={2501.15830},
      archivePrefix={arXiv},
      primaryClass={cs.RO},
}

@misc{CogACT_2024,
      title={CogACT: A Foundational Vision-Language-Action Model for Synergizing Cognition and Action in Robotic Manipulation}, 
      author={Qixiu Li and Yaobo Liang and Zeyu Wang and Lin Luo and Xi Chen and Mozheng Liao and Fangyun Wei and Yu Deng and Sicheng Xu and Yizhong Zhang and Xiaofan Wang and Bei Liu and Jianlong Fu and Jianmin Bao and Dong Chen and Yuanchun Shi and Jiaolong Yang and Baining Guo},
      year={2024},
      eprint={2411.19650},
      archivePrefix={arXiv},
      primaryClass={cs.RO},
}

@misc{VideoVLA_2025,
      title={VideoVLA: Video Generators Can Be Generalizable Robot Manipulators}, 
      author={Yichao Shen and Fangyun Wei and Zhiying Du and Yaobo Liang and Yan Lu and Jiaolong Yang and Nanning Zheng and Baining Guo},
      year={2025},
      eprint={2512.06963},
      archivePrefix={arXiv},
      primaryClass={cs.RO},
}

@misc{GR00T_N1.6,
  title        = {GR00T N1.6: An Improved Open Foundation Model for Generalist Humanoid Robots},
  author       = {GEAR Team and Allison Azzolini and Johan Bjorck and Valts Blukis and Fernando Castañeda and Rahul Chand and others},
  howpublished = {\url{https://research.nvidia.com/labs/gear/gr00t-n1_6/}},
  year         = {2025},
  month        = {December}
}

@misc{FAST_25,
      title={FAST: Efficient Action Tokenization for Vision-Language-Action Models}, 
      author={Karl Pertsch and Kyle Stachowicz and Brian Ichter and Danny Driess and Suraj Nair and Quan Vuong and Oier Mees and Chelsea Finn and Sergey Levine},
      year={2025},
      eprint={2501.09747},
      archivePrefix={arXiv},
      primaryClass={cs.RO},
}

@misc{InternVLA_M1_25,
      title={InternVLA-M1: A Spatially Guided Vision-Language-Action Framework for Generalist Robot Policy}, 
      author={Xinyi Chen and Yilun Chen and Yanwei Fu and Ning Gao and Jiaya Jia and Weiyang Jin and Hao Li and Yao Mu and Jiangmiao Pang and Yu Qiao and Yang Tian and Bin Wang and Bolun Wang and Fangjing Wang and Hanqing Wang and Tai Wang and Ziqin Wang and Xueyuan Wei and Chao Wu and Shuai Yang and Jinhui Ye and Junqiu Yu and Jia Zeng and Jingjing Zhang and Jinyu Zhang and Shi Zhang and Feng Zheng and Bowen Zhou and Yangkun Zhu},
      year={2025},
      eprint={2510.13778},
      archivePrefix={arXiv},
      primaryClass={cs.RO},
}

@misc{ThinkAct_25,
      title={ThinkAct: Vision-Language-Action Reasoning via Reinforced Visual Latent Planning}, 
      author={Chi-Pin Huang and Yueh-Hua Wu and Min-Hung Chen and Yu-Chiang Frank Wang and Fu-En Yang},
      year={2025},
      eprint={2507.16815},
      archivePrefix={arXiv},
      primaryClass={cs.CV},
}

@misc{F1_25,
      title={F1: A Vision-Language-Action Model Bridging Understanding and Generation to Actions}, 
      author={Qi Lv and Weijie Kong and Hao Li and Jia Zeng and Zherui Qiu and Delin Qu and Haoming Song and Qizhi Chen and Xiang Deng and Jiangmiao Pang},
      year={2025},
      eprint={2509.06951},
      archivePrefix={arXiv},
      primaryClass={cs.RO},
      url={https://arxiv.org/abs/2509.06951}, 
}

@inproceedings{deepspeed_2020,
  author = {Rajbhandari, Samyam and Rasley, Jeff and Ruwase, Olatunji and He, Yuxiong},
  title = {ZeRO: memory optimizations toward training trillion parameter models},
  year = {2020},
  isbn = {9781728199986},
  publisher = {IEEE Press},
  booktitle = {Proceedings of the International Conference for High Performance Computing, Networking, Storage and Analysis},
  articleno = {20},
  numpages = {16},
  location = {Atlanta, Georgia},
  series = {SC '20}
}

@misc{PhysBrain_25,
      title={PhysBrain: Human Egocentric Data as a Bridge from Vision Language Models to Physical Intelligence}, 
      author={Xiaopeng Lin and Shijie Lian and Bin Yu and Ruoqi Yang and Changti Wu and Yuzhuo Miao and Yurun Jin and Yukun Shi and Cong Huang and Bojun Cheng and Kai Chen},
      year={2025},
      eprint={2512.16793},
      archivePrefix={arXiv},
      primaryClass={cs.RO},
}

@article{libero,
  title={LIBERO: Benchmarking Knowledge Transfer for Lifelong Robot Learning},
  author={Liu, Bo and Zhu, Yifeng and Gao, Chongkai and Feng, Yihao and Liu, Qiang and Zhu, Yuke and Stone, Peter},
  journal={arXiv preprint arXiv:2306.03310},
  year={2023}
}

@misc{CoT-VLA,
      title={CoT-VLA: Visual Chain-of-Thought Reasoning for Vision-Language-Action Models}, 
      author={Qingqing Zhao and Yao Lu and Moo Jin Kim and Zipeng Fu and Zhuoyang Zhang and Yecheng Wu and Zhaoshuo Li and Qianli Ma and Song Han and Chelsea Finn and Ankur Handa and Ming-Yu Liu and Donglai Xiang and Gordon Wetzstein and Tsung-Yi Lin},
      year={2025},
      eprint={2503.22020},
      archivePrefix={arXiv},
      primaryClass={cs.CV},
}

@misc{LangForce,
      title={LangForce: Bayesian Decomposition of Vision Language Action Models via Latent Action Queries}, 
      author={Shijie Lian and Bin Yu and Xiaopeng Lin and Laurence T. Yang and Zhaolong Shen and Changti Wu and Yuzhuo Miao and Cong Huang and Kai Chen},
      year={2026},
      eprint={2601.15197},
      archivePrefix={arXiv},
      primaryClass={cs.AI},
}

@misc{ABot-M0_26,
      title={ABot-M0: VLA Foundation Model for Robotic Manipulation with Action Manifold Learning}, 
      author={Yandan Yang and Shuang Zeng and Tong Lin and Xinyuan Chang and Dekang Qi and Junjin Xiao and Haoyun Liu and Ronghan Chen and Yuzhi Chen and Dongjie Huo and Feng Xiong and Xing Wei and Zhiheng Ma and Mu Xu},
      year={2026},
      eprint={2602.11236},
      archivePrefix={arXiv},
      primaryClass={cs.CV},
}

@misc{TwinBrainVLA,
      title={TwinBrainVLA: Unleashing the Potential of Generalist VLMs for Embodied Tasks via Asymmetric Mixture-of-Transformers}, 
      author={Bin Yu and Shijie Lian and Xiaopeng Lin and Yuliang Wei and Zhaolong Shen and Changti Wu and Yuzhuo Miao and Xinming Wang and Bailing Wang and Cong Huang and Kai Chen},
      year={2026},
      eprint={2601.14133},
      archivePrefix={arXiv},
      primaryClass={cs.RO},
}

@misc{VLA-JEPA_26,
      title={VLA-JEPA: Enhancing Vision-Language-Action Model with Latent World Model}, 
      author={Jingwen Sun and Wenyao Zhang and Zekun Qi and Shaojie Ren and Zezhi Liu and Hanxin Zhu and Guangzhong Sun and Xin Jin and Zhibo Chen},
      year={2026},
      eprint={2602.10098},
      archivePrefix={arXiv},
      primaryClass={cs.RO},
}

@inproceedings{ReMix_24,
  title={ReMix: Optimizing Data Mixtures for Large Scale Imitation Learning},
  author={Joey Hejna and Chethan Anand Bhateja and Yichen Jiang and Karl Pertsch and Dorsa Sadigh},
  booktitle={8th Annual Conference on Robot Learning},
  year={2024},
}

@misc{demonstration-info_25,
      title={Robot Data Curation with Mutual Information Estimators}, 
      author={Joey Hejna and Suvir Mirchandani and Ashwin Balakrishna and Annie Xie and Ayzaan Wahid and Jonathan Tompson and Pannag Sanketi and Dhruv Shah and Coline Devin and Dorsa Sadigh},
      year={2025},
      eprint={2502.08623},
      archivePrefix={arXiv},
      primaryClass={cs.RO},
}

@inproceedings{SCIZOR_26,
  title={SCIZOR: Self-Supervised Data Curation for Large-Scale Imitation Learning},
  author={Zhang, Yu and Xie, Yuqi and Liu, Huihan and Shah, Rutav and Wan, Michael and Fan, Linxi and Zhu, Yuke},
  booktitle={IEEE International Conference on Robotics and Automation (ICRA)},
  year={2026}
}

@misc{TGM-VLA_26,
      title={TGM-VLA: Task-Guided Mixup for Sampling-Efficient and Robust Robotic Manipulation}, 
      author={Fanqi Pu and Lei Jiang and Wenming Yang},
      year={2026},
      eprint={2603.00615},
      archivePrefix={arXiv},
      primaryClass={cs.RO},
}

\newpage

\appendix

\section{Additional Implementation Details}
\label{app:implementation}

\subsection{Pruning Cache}
\label{app:cache}

\methodname{} stores trajectory-level pruning results in a cache containing the original importance scores and the retained indices for each configured ratio. The cache supports reuse across experiments as long as the importance and pruning configurations remain compatible. During distributed training, cache construction can be restricted to rank zero and loaded by other workers after synchronization.

\subsection{Frame-Score Preprocessing}
\label{app:score_preprocessing}

For VAC, we use a DINOv2 visual encoder and extract visual features from at most 16 sparsely sampled video frames per trajectory before interpolating VAC scores back to the original trajectory length. The implementation records frame extraction failures and trajectories without usable visual features so that unreliable preprocessing runs can be identified before training.

For GMM-TPI, we fit the progress prior separately for each benchmark using 5\% of the corresponding training trajectories. The annotation records only the normalized progress locations of manipulation-critical stage centers, such as alignment, grasping, and release; it does not provide action labels, success labels, or per-frame supervision to the policy. We fit a three-component one-dimensional GMM over these progress values and normalize the resulting density within each trajectory before adding it to the frame-importance score. This prior is used only during offline frame-score preprocessing. The VLA policy, training loss, and evaluation protocol do not access these annotations. When such annotations are unavailable, we use the dataset-agnostic Gaussian prior described in Section~\ref{sec:method}.

\subsection{Main Training Schedule}
\label{app:training_schedule}

For the main experiments, the active compressed view uses a retention ratio of $r=0.2$, corresponding to an 80\% frame pruning ratio per trajectory. Training alternates between five mini-batches from this pruned view and one mini-batch from the full-frame view with $r=1.0$. The full-frame mini-batch is used only as a periodic context anchor; evaluation is performed with the standard policy inference procedure and does not require frame pruning.

Because the three benchmarks use different training datasets and established evaluation recipes, we follow the commonly used training budgets for each benchmark rather than forcing a single step count across all settings. For RoboCasa-GR1, we train on the corresponding expert demonstration data for 100K optimization steps. For SimplerEnv, we train on the BridgeV2 dataset for 60K optimization steps. For LIBERO, we train on expert teleoperation demonstrations for 30K optimization steps.

\section{Full RoboCasa-GR1 Results}
\label{app:robocasa_full}

\begin{table*}[!t]
    \centering
    \small
    \renewcommand{\arraystretch}{1.4} 
    \setlength{\tabcolsep}{1.6pt} 

    \caption{
      \textbf{Results of evaluating the VLA models with the GR1 robot in the RoboCasa-GR1 Tabletop simulation environment}. The results for Isaac-GR00T N1.5 and Isaac-GR00T N1.6 are sourced from the official Isaac-GR00T GitHub repository~\cite{GR00T_25_arxiv}. We highlight the best results in \textbf{bold} and the second-best results with \underline{underline}.
    }
    \begin{adjustbox}{width=\textwidth}
    \rowcolors{2}{gray!15}{white}
    \begin{tabular}{l c c c c c c c}
        \toprule
        {Task} & 
        {\scriptsize \makecell{\textbf{GR00T N1.5}}} & 
        {\scriptsize \makecell{\textbf{GR00T N1.6}}} & 
        {\scriptsize \makecell{\textbf{VP-VLA}}} & 
        {\scriptsize \makecell{\textbf{TwinBrainVLA}}} & 
        {\scriptsize \makecell{\textbf{PhysBrain}}} & 
        {\scriptsize \makecell{\textbf{LangForce}}} &
        {\scriptsize \makecell{\textbf{FrameSkip}}} \\
        \midrule
        PnP Bottle To Cabinet Close                                             & 54.0 & 51.5 & 54.0 & 74.0 & 74.0 & 72.0 & 74.0 \\
        PnP Can To Drawer Close                                                 & 50.0 & 13.0 & 72.0 & 72.0 & 68.0 & 78.0 & 82.0 \\
        PnP Cup To Drawer Close                                                 & 38.0 &  8.5 & 44.0 & 52.0 & 42.0 & 46.0 & 46.0 \\
        PnP Milk To Microwave Close                                             & 60.0 & 14.0 & 74.0 & 60.0 & 54.0 & 56.0 & 64.0 \\
        PnP Potato To Microwave Close                                           & 32.0 & 41.5 & 34.0 & 36.0 & 24.0 & 36.0 & 46.0 \\
        PnP Wine To Cabinet Close                                               & 38.0 & 16.5 & 48.0 & 46.0 & 54.0 & 46.0 & 76.0 \\
        \midrule
        \rowcolor{gray!20}\textbf{PnP * to * Close (Avg)}                       & 45.3 & 24.2 & 54.3 & \underline{56.7} & 52.7 & 55.7 & \textbf{63.7} \\
        \midrule
        PnP Novel From Cuttingboard To Basket                                   & 38.0 & 58.0 & 66.0 & 62.0 & 62.0 & 66.0 & 58.0 \\
        PnP Novel From Cuttingboard To Cardboardbox                             & 46.0 & 46.5 & 54.0 & 46.0 & 44.0 & 40.0 & 58.0 \\
        PnP Novel From Cuttingboard To Pan                                      & 58.0 & 68.5 & 74.0 & 70.0 & 56.0 & 68.0 & 70.0 \\
        PnP Novel From Cuttingboard To Pot                                      & 62.0 & 65.0 & 54.0 & 66.0 & 58.0 & 48.0 & 66.0 \\
        PnP Novel From Cuttingboard To Tieredbasket                             & 28.0 & 46.5 & 56.0 & 52.0 & 40.0 & 44.0 & 54.0 \\
        \midrule
        \rowcolor{gray!20}\textbf{PnP Novel From Cuttingboard To * (Avg)}       & 46.4 & 56.9 & \underline{60.8} & 59.2 & 52.0 & 53.2 & \textbf{61.2} \\
        \midrule
        PnP Novel From Placemat To Basket                                       & 30.0 & 58.5 & 48.0 & 30.0 & 42.0 & 54.0 & 52.0 \\
        PnP Novel From Placemat To Bowl                                         & 60.0 & 57.5 & 74.0 & 54.0 & 56.0 & 62.0 & 66.0 \\
        PnP Novel From Placemat To Plate                                        & 56.0 & 63.0 & 70.0 & 64.0 & 80.0 & 52.0 & 66.0 \\
        PnP Novel From Placemat To Tieredshelf                                  & 36.0 & 28.5 & 26.0 & 38.0 & 14.0 & 24.0 & 30.0 \\
        \midrule
        \rowcolor{gray!20}\textbf{PnP Novel From Placemat To * (Avg)}           & 45.5 & 51.9 & \textbf{54.5} & 46.5 & 48.0 & 48.0 & \underline{53.5} \\
        \midrule
        PnP Novel From Tray To Cardboardbox                                     & 52.0 & 51.5 & 44.0 & 46.0 & 40.0 & 50.0 & 54.0 \\
        PnP Novel From Tray To Plate                                            & 48.0 & 71.0 & 66.0 & 72.0 & 66.0 & 58.0 & 62.0 \\
        PnP Novel From Tray To Pot                                              & 60.0 & 64.5 & 38.0 & 56.0 & 52.0 & 62.0 & 66.0 \\
        PnP Novel From Tray To Tieredbasket                                     & 52.0 & 57.0 & 58.0 & 46.0 & 50.0 & 44.0 & 66.0 \\
        PnP Novel From Tray To Tieredshelf                                      & 32.0 & 31.5 & 24.0 & 28.0 & 22.0 & 22.0 & 40.0 \\
        \midrule
        \rowcolor{gray!20}\textbf{PnP Novel From Tray To * (Avg)}               & 48.8 & \underline{55.1} & 46.0 & 49.6 & 46.0 & 47.2 & \textbf{57.6} \\
        \midrule
        PnP Novel From Plate To Bowl                                            & 58.0 & 57.0 & 52.0 & 60.0 & 54.0 & 54.0 & 58.0 \\
        PnP Novel From Plate To Cardboardbox                                    & 44.0 & 43.5 & 44.0 & 46.0 & 50.0 & 48.0 & 46.0 \\
        PnP Novel From Plate To Pan                                             & 60.0 & 51.0 & 56.0 & 56.0 & 68.0 & 54.0 & 62.0 \\
        PnP Novel From Plate To Plate                                           & 64.0 & 78.7 & 62.0 & 66.0 & 78.0 & 78.0 & 72.0 \\
        \midrule
        \rowcolor{gray!20}\textbf{PnP Novel From Plate To * (Avg)}              & 56.5 & 57.6 & 53.5 & 57.0 & \textbf{62.5} & 58.5 & \underline{59.5} \\
        \midrule
        \rowcolor{gray!30} 
        \textbf{Average}                                                        & 48.2 & 47.6 & 53.8 & \underline{54.6} & 50.0 & 52.6 & \textbf{59.5} \\
        \bottomrule
    \end{tabular}
    \end{adjustbox}
    \label{tab:robocasa_main_tab}
\end{table*}






\end{document}